# A Systematic Analysis of Live Intraoperative Surgical Expert Narrations

Authors:

*Arda Mamur [1,3], Matthias Seibold [1,3], Christoph Laux [2,3], Florian Wanivenhaus [2,3] & Philipp Fürnstahl [1,3]*

Affiliations:

*[1] Research in Orthopedic Computer Science (ROCS), Balgrist University Hospital, Zurich, Switzerland*
*[2] University Spine Center Zurich, Balgrist University Hospital, Zurich, Switzerland*
*[3] University of Zurich, Zurich, Switzerland*

Keywords:

*weak supervision, self-supervised, surgical video understanding, intraoperative narration, surgical workflow*

Key information:

1. **Research question:** Manual annotation remains the principal bottleneck in surgical video understanding; live intraoperative expert surgeon narration is a scalable alternative that requires no manual annotation, but what information it contains has not been characterized. This study investigates which supervisory signals are contained in continuous live intraoperative narrations (e.g., surgical actions, workflow phases, instrument usage, anatomy, and surgical intent), what sources of noise are introduced, and how intraoperative narrations can be processed to leverage them for safe downstream applications.
2. **Findings:** Across 13 endoscopic spine decompression procedures, live narration recorded from two senior spine surgeons proved a dense but noisy source of procedural information: 98.5% of retained cues referenced surgical instruments (60.7%), anatomy (48.0%), or actions (57.1%), and 39% (472 of 1,201) were directives, largely instrument requests and commands that were consistently anticipatory, spoken before the event they call for. The noise is systematic and measurable: 18% of retained cues carried domain-term transcription corruption, and 57.6% of raw speech was non-procedural and removed by filtering.
3. **Meaning:** Live intraoperative narration provides dense supervisory signals describing instruments, anatomy, surgical actions, and team coordination, and unlike the retrospective lecture narration used in related works it is generated concurrently with the procedure and is therefore temporally coupled to operative events. At the same time, narration may precede or follow the corresponding action, creating a temporal misalignment that must be characterized and modelled before safe downstream use.

MANUSCRIPT

## Introduction

Surgical video understanding and workflow recognition have advanced with domain foundation models,[1] yet supervision still depends on manual per-frame annotation which remains the principal bottleneck.[2] In general domains, speech aligned with video can replace manual labels,[3-5] and recent surgical work applies this using post-hoc lecture narration.[6-8] Live intraoperative narration, produced by the surgeon while operating, is a more authentic and scalable signal, yet uncharacterized. In this study, we investigate whether intraoperative narrations can be reduced to procedural cues that are usable for surgical context understanding, and characterize their density, composition, noise, and temporal alignment.

## Material and methods

Thirteen endoscopic lumbar spine decompression procedures by two senior spine surgeons at one center were analysed: endoscopic video with time-synchronized close-mic surgeon audio in German, 13.2 h total, without patient-identifying data.

**Characterization pipeline.** Established components requiring no manual annotation were chained (Figure 1): word-level automatic speech recognition;[9] deterministic normalization of recurrent surgical-term transcription errors; fusion of pause-bounded segments into utterances; large-language-model German-to-English translation with surgical-context prompting. Each utterance was classified as procedural or non-procedural; each procedural utterance became a *cue*, a time-stamped record labelled with the surgical entities it references (anatomy, instruments, actions, tissue states, intents; non-exclusive) and with its speech act, observational or directive.

**Analysis.** Outcomes are descriptive, quantified per procedure and pooled: cue density (cues per minute of surgery); occurrence frequency of each entity category among retained cues; share of speech discarded as non-procedural; domain-term corruption rate; offset between directive cues and the corresponding on-screen event; and per-stage cue retention (Figure 2).

## Results

Surgeon speech occupied 29.9% of recording time; 3,280 ASR segments were fused into 3,171 utterances, of which 1,201 (37.9%) were retained as procedural cues (729 observational, 472 directive), a density of 1.5 cues per minute of recorded time; 57.6% were removed as non-procedural, 2.0% were unintelligible and 2.5% unclassifiable (Figure 2). Retained cues referenced instruments in 60.7%, anatomical structures in 48.0% and surgical actions in 57.1% (non-exclusive; 98.5% carried at least one). Domain-term corruption affected 18.0% of retained cues (216 of 1,201), over half of all corrections falling on a single instrument term. All instrument-request cues (467) were manually verified against the video; the request preceded the instrument's appearance in the field in every case, by a median of 8.2 s (Figure 3).

## Discussion and Conclusion

Narration is therefore dense enough to supply the categories a workflow model requires, and it can be captured reliably despite operating-room acoustics; reliable capture under these conditions is itself a finding.

Unlike post-hoc lecture narration,[6-8] live narration is produced while a team is being addressed and therefore contains directives. In our dataset, 39% of the cues have been identified as instrument requests and commands that precede the events they name. This register carries surgical intent and workflow progression absent from label-derived supervision, but it is also misaligned. A request aligned to the moment it is spoken pairs speech with the wrong visual state (Figure 3); because temporal and action understanding is a known weakness of video-language models,[10] such an association would be learned, not detected. Live narration is thus a scalable, annotation-free supervisory signal whose temporal displacement must be modelled before downstream use.

**Limitations.** The corpus is single-center, single-specialty and single-language; the error families reported may not transfer. Temporal displacement was quantified for instrument-request cues only. Cohort-wide offset estimation and multi-center replication are the next steps.

References

1. Wang Z, Liu C, Zhang S, Dou Q. Foundation model for endoscopy video analysis via large-scale self-supervised pre-train. In: *International Conference on Medical Image Computing and Computer-Assisted Intervention*. Springer; 2023:101-111.
2. Seenivasan L, Islam M, Krishna AK, Ren H. Surgical-VQA: visual question answering in surgical scenes using transformer. In: *International Conference on Medical Image Computing and Computer-Assisted Intervention*. Springer; 2022:33-43.
3. Miech A, Zhukov D, Alayrac JB, Tapaswi M, Laptev I, Sivic J. HowTo100M: learning a text-video embedding by watching hundred million narrated video clips. In: *Proceedings of the IEEE/CVF International Conference on Computer Vision*. 2019:2630-2640.
4. Radford A, Kim JW, Hallacy C, et al. Learning transferable visual models from natural language supervision. In: *International Conference on Machine Learning*. PMLR; 2021:8748-8763.
5. van den Oord A, Li Y, Vinyals O. Representation learning with contrastive predictive coding. *arXiv*. Preprint posted July 11, 2018. arXiv:1807.03748
6. Yuan K, Srivastav V, Yu T, et al. Learning multi-modal representations by watching hundreds of surgical video lectures. *Med Image Anal*. 2025;105:103644.
7. Yuan K, Srivastav V, Navab N, Padoy N. HecVL: hierarchical video-language pretraining for zero-shot surgical phase recognition. In: *International Conference on Medical Image Computing and Computer-Assisted Intervention*. Springer; 2024:306-316.
8. Yuan K, Srivastav V, Navab N, Padoy N. Procedure-aware surgical video-language pretraining with hierarchical knowledge augmentation. *Adv Neural Inf Process Syst*. 2024;37:122952-122983.
9. Bain M, Huh J, Han T, Zisserman A. WhisperX: time-accurate speech transcription of long-form audio. In: *Proceedings of INTERSPEECH 2023*. 2023.
10. Ma W, Li K, Jiang Z, et al. Rethinking video-text understanding: retrieval from counterfactually augmented data. In: *European Conference on Computer Vision*. Springer; 2024:254-269.

Disclosures

The authors declare no conflict of interest related to this work. This study received no funding and no industrial support.

*APPENDIX*

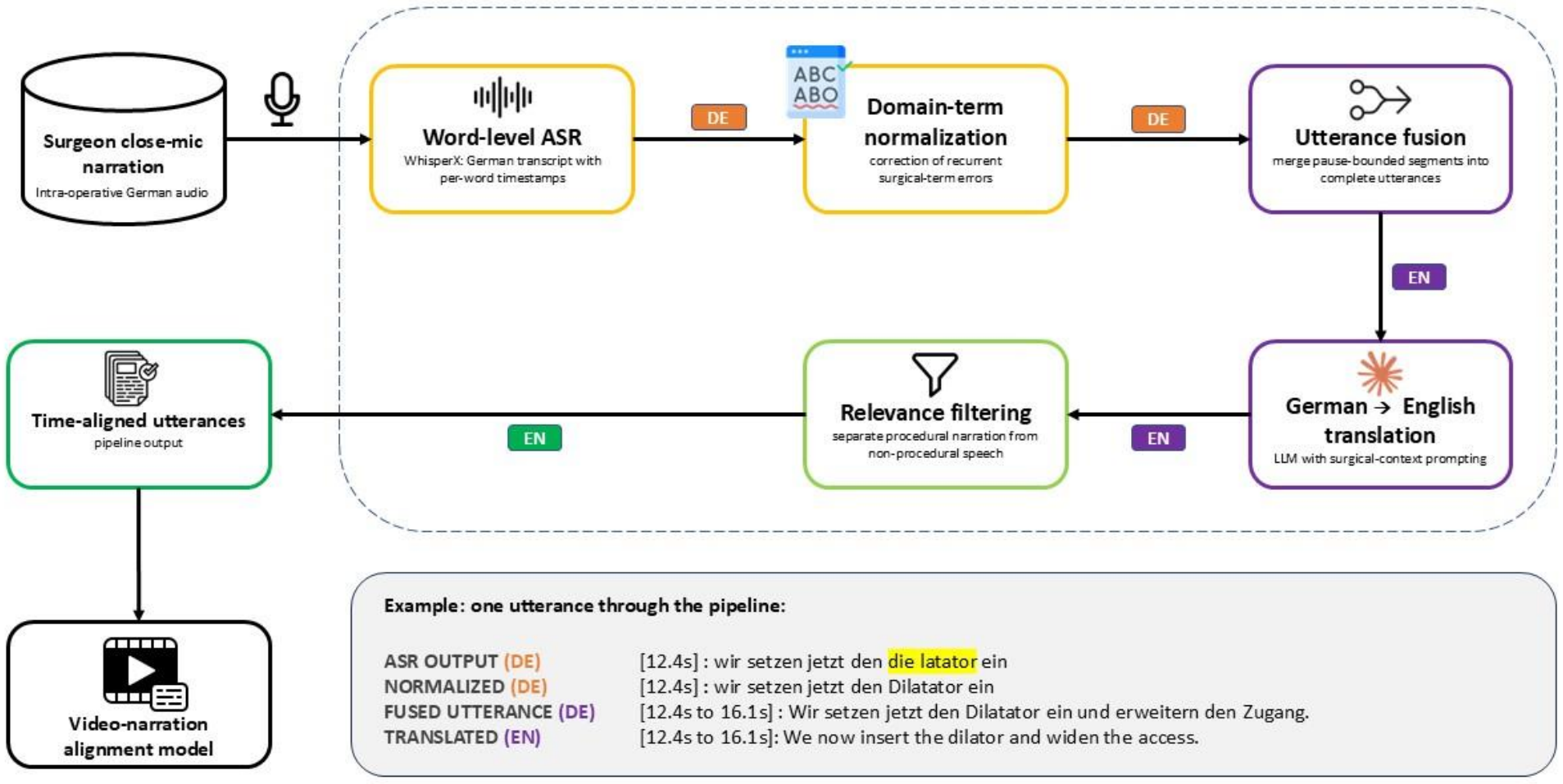

**Figure 1:** Annotation-free pipeline for characterizing intraoperative surgeon narration. WhisperX transcribes close-mic German audio into word-level, time-stamped text (1); recurrent surgical-term errors are corrected deterministically (2); pause-bounded segments are fused into utterances (3); a large language model translates each utterance into English with surgical-context prompting (Claude Opus 4.8) (4); and relevance filtering separates procedural from non-procedural speech (5). Edge labels give the stream language (DE/EN); the inset traces one utterance through all stages.

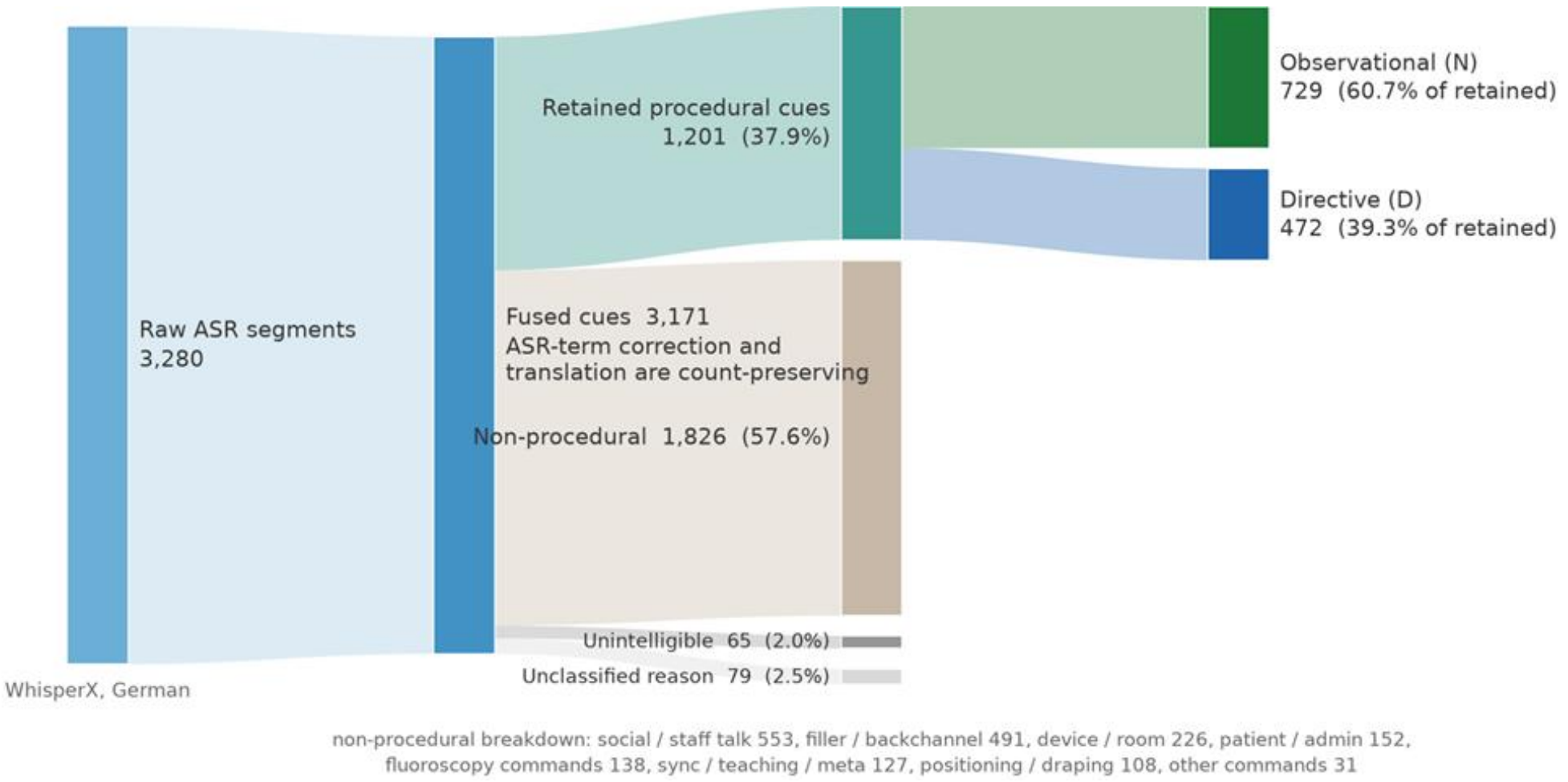

**Figure 2:** Per-stage cue-retention funnel (13 procedures). WhisperX transcription yields 3,280 raw ASR segments, fused into 3,171 utterance-level cues. Relevance filtering retains 1,201 procedural cues (37.9%; 729 observational, 472 directive) and removes 1,826 non-procedural cues (57.6%), 65 unintelligible (2.0%), and 79 unclassified (2.5%); the breakdown of non-procedural speech is shown in the diagram.

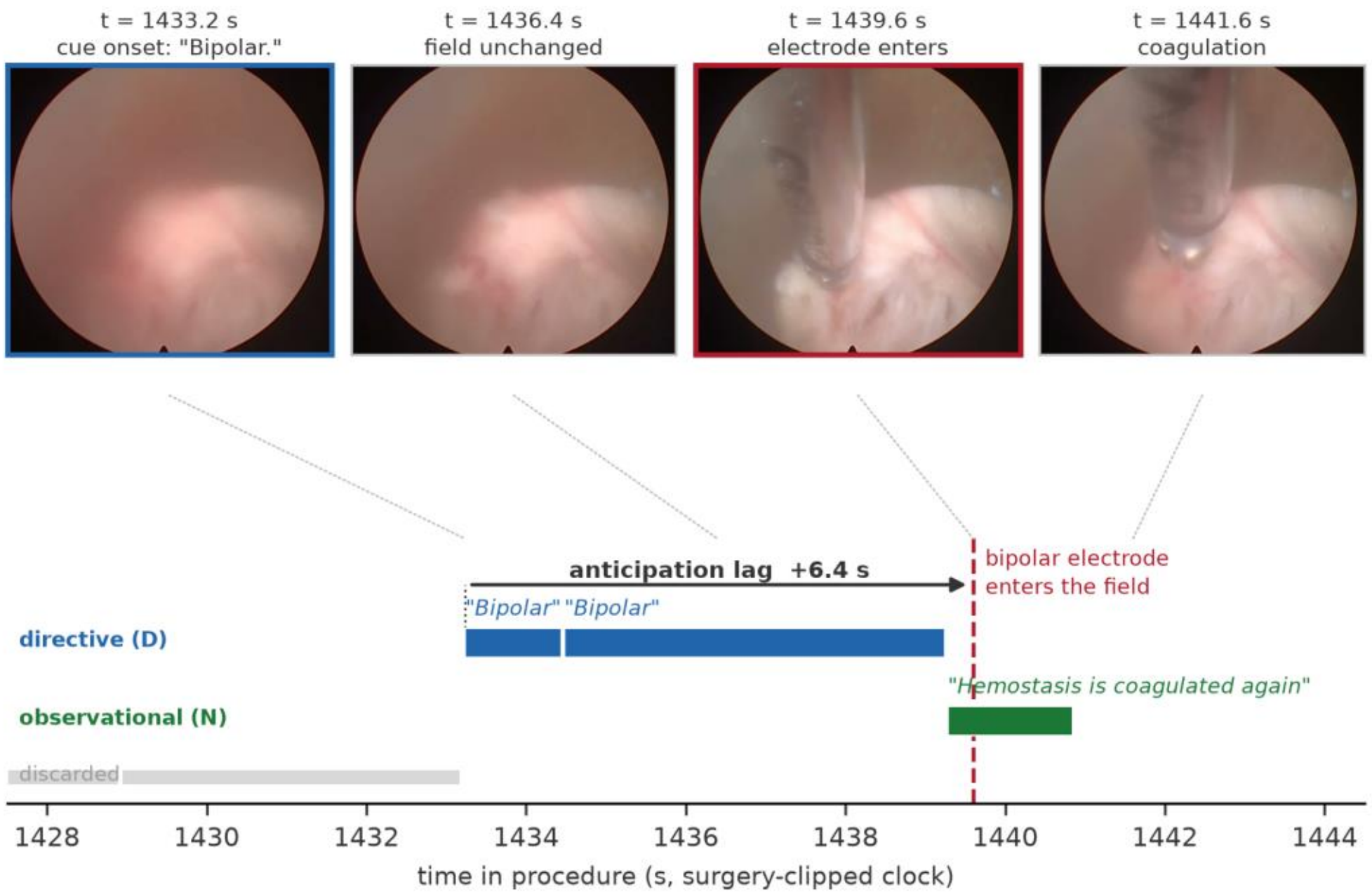

**Figure 3:** Anticipatory directive example. The instrument request ("Bipolar") is spoken at t = 1433.2 s while the field is unchanged; the bipolar electrode enters the field 6.4 s later. Aligning the directive to its utterance time would pair speech with a frame that does not yet contain the instrument.